\documentclass[letterpaper, journal]{IEEEtran}
\usepackage{amsmath,amsfonts}
\usepackage{algorithmic}
\usepackage{algorithm}
\usepackage{array}
\usepackage{textcomp}
\usepackage{stfloats}
\usepackage{url}
\usepackage{verbatim}
\usepackage{graphicx}
\usepackage{cite}
\usepackage{amsthm}
\usepackage{subfigure}
\usepackage{enumitem}
\usepackage{multicol}
\usepackage{multirow}
\usepackage{tabularx}
\usepackage{caption}
\usepackage{hyperref}

\newtheorem{definition}{Definition} 

\begin{document}


\title{Incremental Causal Graph Learning for Online Cyberattack Detection in Cyber-Physical Infrastructures}

\author{\IEEEauthorblockN{Arun Vignesh Malarkkan, Dongjie Wang, Haoyue Bai, and Yanjie Fu}

\IEEEauthorblockA{A.V. Malarkkan, H. Bai, and Y. Fu are with the School of Computing and Augmented Intelligence, Arizona State University, Tempe, Arizona, USA. \\
E-mail: \{arun.malarkkan, haoyuba, yanjie.fu\}@asu.edu}

\IEEEauthorblockA{D. Wang is with the Institute for Information Sciences, University of Kansas, Lawrence, Kansas, USA. \\
E-mail: wangdongjie@ku.edu}
}





\maketitle

\begin{abstract}
The escalating threat of cyberattacks on real-time critical infrastructures poses significant risks to public safety, necessitating detection methods that can effectively capture complex system interdependencies and adapt to evolving attack patterns. Traditional real-time anomaly detection techniques often produce excessive false positives due to their statistical sensitivity to high data variability and class imbalance. To address these limitations, recent research has explored modeling causal relationships among system components. However, prior work predominantly focuses on offline causal graph-based approaches that require static historical data and fail to generalize to real-time settings. These methods are fundamentally constrained by:  
(1) their inability to adapt to dynamic shifts in data distribution without retraining, and  
(2) the risk of catastrophic forgetting when lacking timely supervision in live systems.
To overcome these challenges, we propose \textbf{INCADET}, a novel framework for \textit{incremental causal graph learning} tailored to real-time cyberattack detection. INCADET dynamically captures evolving system behavior by incrementally updating causal graphs across streaming time windows. The framework comprises three modules:  
\textbf{1) Early Symptom Detection:} Detects transitions in system status using divergence in edge-weight distributions across sequential causal graphs.  
\textbf{2) Incremental Causal Graph Learning:} Leverages experience replay and edge reinforcement to continually refine causal structures while preserving prior knowledge.  
\textbf{3) Causal Graph Classification:} Employs Graph Convolutional Networks (GCNs) to classify system status using the learned causal graphs.
Extensive experiments on real-world critical infrastructure datasets demonstrate that INCADET achieves superior accuracy, robustness, and adaptability compared to both static causal and deep temporal baselines in evolving attack scenarios.
\end{abstract}


\section{Introduction}
In real-world critical public infrastructures, adversarial cyberattacks emerge incrementally, evolving from subtle data perturbations to complex intrusions that trigger delayed, cascading disruptions across interconnected nodes, complicating detection and mitigation. The 2021 US Colonial Fuel Pipeline Ransomware Attack is one such instance that exemplified an incremental cyber threat, where a breach triggered cascading disruptions in the fuel monitoring and alarm systems, amplifying economic impact over time. Consequently, detection and mitigation of such cyberattacks can be generalized as Incremental Graph-based Anomaly Detection in a Multivariate Time-Series setting \cite{Pinto2024}. Lately, the research on Incremental or Continual Graphs \cite{6977167, 10.14778/3579075.3579088, Liu_2023, 10.1145/3580305.3599392} 
 and offline causal graph-based detection  \cite{10.1145/3627673.3680096, febrinanto2023entropy, malarkkan2025rethinkingspatiotemporalanomalydetection} have emerged as a promising approach to effectively represent streaming data, enabling real-time threat detection and distinguishing time-lagged anomalies. An incremental causal graph anomaly detection approach can address some critical issues when it comes to real-world cyberattack detection use cases. For example, 1) Real-time adaptation: Cyber-physical systems are dynamic, experiencing constant changes in operational conditions with respect to the interconnected nodes. 
2) Evolving Attack Patterns: Cyberattacks are not static; adversaries constantly adapt their techniques with new intrusion mechanisms. 
3) Partial Observability in real-time large systems: The statistical observability of the entire inter-connected system evolves and is unstable over time. Causal Graphs identify true cause-effect relationships, disentangling them from noisy correlation-based edges and maintaining stability across data distributions.
Thus, developing an adaptive detection framework for dynamic and temporally dependent multivariate systems is vital to ensuring infrastructure resilience against evolving cyber threats.
 
\begin{figure}[t]
  \centering
  \includegraphics[width=0.99\linewidth]{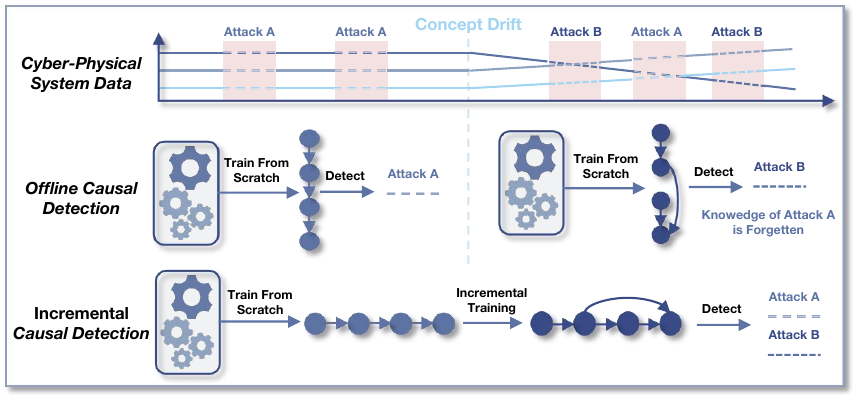}
  \caption{\textbf{Offline Causal Graph-based Cyberattack Detection Workflow vs. Proposed Incremental Causal Graph-based Cyberattack Detection Workflow}}
  \vspace{-0.5cm}
\end{figure}

 However, there are two major research challenges in implementing an Incremental Causal Graph Framework:
 \textbf{Challenge 1: Catastrophic Forgetting and Dynamic Concept Drift.} In constructing an Incremental Causal Graph framework, the graphs keep evolving as the model learns from the new batch of data in a dynamic setting. A key challenge is knowledge forgetting from previous causal relationships in data segments, particularly during concept drift. This is called Catastrophic Forgetting. It prevents the framework from adapting to evolving important causal feature relationships during concept drift, hindering performance.
 \textbf{Challenge 2: Efficient updation of Incremental Causal Graphs.} The construction of an efficient incremental causal graph framework to accurately represent an attack event relies on the assumption that only prior vulnerable nodes and their direct causal descendants experience impact. However, this might not hold true in real-world scenarios, where complex, multi-hop causal pathways can propagate attacks. At the same time, maintaining knowledge of all past attack events and their corresponding causal structures in large-scale systems incurs substantial memory and computational overhead. Balancing memory usage and incremental causal structure accuracy over time is the key challenge.

Earlier incremental time-series anomaly detection relied on statistical optimization like SVM kernels \cite{laskov2006incremental, yi2011incremental} and density-based clustering \cite{ren2008using, burbeck2007adaptive}, but these methods were sensitive to initial conditions and required dynamic parameter tuning, leading to catastrophic forgetting. To overcome these limitations, recent research on incremental graph-based anomaly detection \cite{gao2025dynamic, 10577591, 6977167, 10.14778/3579075.3579088, Liu_2023, 10.1145/3580305.3599392} recognized their potential to handle dynamic data streams and accurately represent evolving system behaviors using Graph Convolutional Networks. These approaches however faced scalability challenges, particularly memory and computational efficiency when incorporating past knowledge.
Also, with complex interconnections and data imbalance, real-world graph learning suffers from spurious correlational connections, reducing detection accuracy. Recent advances in Causal Graph Learning \cite{10.1145/3627673.3680096, febrinanto2023entropy} address these challenges by identifying true cause-effect relationships, disentangling them from noisy correlation-based edges, and maintaining stability across data distributions.
However, in a streaming multivariate setting, offline causal anomaly detection is hindered by spurious causal connections, missed delayed temporal dependencies, and the need for costly retraining on large historical data. These challenges motivate us to ask the following questions.
1) Is it possible to model a robust causal graph-based anomaly detection for dynamic, evolving real-world systems? 
2) Can we deploy an online causal learning-based anomaly detection model adapting to unseen events and constant drifts?
3) Can we build a resource-efficient approach without the need for retraining the causal model on the entire dataset?

\textbf{Our research perspective is to combine the two paradigms of Incremental Graphs and Causal Graph Learning to address online anomaly detection}. Our perspective is bi-fold: 1) Robust Causal Knowledge Representation: In this work, we highlight that in real-world dynamic systems, correlations can change rapidly, while causal relationships tend to be more stable. With the underlying causal structures, the detection system becomes more robust, allowing the incremental framework to adapt to evolving normal behavior without losing its ability to detect genuine anomalies when the streaming data distribution changes. 2) Memory-efficient and Scalable Detection Framework: We also highlight that memory efficiency allows for the storage of relevant causal relationships and historical data without excessive resource consumption when the scale grows. It allows for the representation and manipulation of continual causal structures without degrading performance.

Inspired by these research insights, we propose INCADET, a novel unsupervised incremental causal graph learning framework for online cyberattack detection in critical infrastructures. Our framework consists of three main modules. 1) Early Symptom Detection Phase; 2) Incremental Causal Graph Construction Phase; 3) Deep Graph Convolutional Network-based Attack Detection Phase. In the first module, we develop an early symptom trigger point detection module based on casual edge weight divergence between the static causal graphs of subsequent time segments. In the second module, two incremental causal graphs corresponding to the system status are constructed. Here, we utilize the prior attack events' knowledge in the Replay Knowledge Buffer to propagate across time segments as Incremental Knowledge. This technique efficiently prevents the model from "Catastrophic Forgetting" of the crucial attack knowledge from the past. Past cyberattack event edges stored in the replay buffer are reweighted by their future occurrence frequency through Causal Edge Reinforcement. It enforces that the causal graph constructed is more robust to perturbations caused by unseen dynamic state changes. The continual construction of the graphs is stopped when the edge distribution converges between the static causal graphs across time segments. In the third module, we use the constructed causal graphs as ground truth into a Deep Graph Convolutional Network to efficiently detect cyberattacks per time-interval segment. To summarize, the main contributions of the paper are as follows.
\begin{enumerate}
    \item \textbf{Research Problem:} We investigate an important research problem of online cyberattack detection in critical infrastructures and propose to address the research challenges with an incremental causal perspective, and have experimentally highlighted why the incremental perspective is better than the existing offline causal-based approaches.
    \item \textbf{Proposed Framework:} We propose INCADET, an unsupervised Incremental Causal Graph Learning Framework, with Early Symptom Detection, Incremental Causal Graph Learning with Replay Buffer and Causal Edge Reinforcement, and Graph Convolutional Network modules.
    \item \textbf{Experimental Evaluation:} Through a series of experiments on real-world datasets, we successfully demonstrate our framework's superior performance in comparison to existing graph-based and offline causal approaches in terms of balancing scalability and efficiency.
\end{enumerate}

\section{Preliminaries}

\begin{definition}
    \textbf{Incremental Graph Construction} is a research paradigm that refers to the process of constructing and continually updating a graph as a new stream of data arrives. The constructed graph is incrementally updated with streaming data, allowing for structural adaptation through node, edge, and attribute changes.
\end{definition}
\begin{definition}
    \textbf{Replay Buffer} is an important technique we have utilized to store crucial attack information from the causal graphs in the past. We use the stored information in the buffer to retain and propagate the historical causal knowledge through the constructed causal graph in the current time interval \cite{balaji2020effectiveness}.
\end{definition}
\begin{definition}
    \textbf{Causal Edge Reinforcement} refers to the re-adjustment of weights of edges based on it causal influence from the past in the continual graph construction. It reinforces key causal edges representing past attacks while weakening spurious edges that disrupt attack knowledge.
\end{definition}

\textbf{Problem Statement}. 
Detecting cyberattacks in critical infrastructures online can be technically formulated as unsupervised online anomaly detection in multivariate time-series data. 
A multivariate time-series sequence of data points is given by
\begin{center}
    $\mathbf{X} = \{x^i_1, x^i_2,...,x^i_{t-1}, x^i_t\}, x^i \in \mathbb{R}^M $
\end{center}
$x^i$ is a $M$ dimensional vector at time $t$ $(t<=T)$ which means, at each timestamp $t$, $M$ features belong to the data point $x$. $X$ are non-static observations that change over time. 
We assume that when a cyberattack event happens in the system, these observations show drastic changes implying the immediate, cascading impact on the system.
We formulate the first step in countering this cyberattack as an unsupervised anomaly detection problem with $\mathbf{X}$ as the training input of streaming data points. 
We define a time-segment window of $k$ seconds such that each window of observations is assigned a system status $y \in \{0,1\}$ representing the statuses "Normal" and "Attack" respectively. Our cyberattack detection framework aims to assign a label $\hat{y} \in \{0,1\}$ to an unseen window of $k$ seconds, determining the system status per time segment.

\section{Methodology}
In this paper, we propose INCADET, an incremental causal graph learning-based framework for online cyberattack detection in critical infrastructures. 
INCADET consists of three closely cooperating stages: early symptom trigger detection, which is responsible for determining the necessity of incremental causal graph learning to achieve timely and valuable detection; incremental causal graph construction, which is responsible for smoothly learning new causal patterns while avoiding catastrophic forgetting; and deep graph convolutional network-based graph classification, which is responsible for reliable and explainable attack detection.
Initially, the early symptom trigger detection stage constructs a static causal structural model of nominal operational behavior from feature-engineered multivariate time series observations. 
Static causal graphs are generated for consecutive time windows to detect early attack symptoms. The divergence between the edge weight distributions of these graphs is then quantified using the Jensen-Shannon Divergence. When this divergence exceeds a predefined threshold, it signals a potential attack precursor, thereby triggering the incremental learning phase. 
Subsequently, the incremental causal graph construction stage constructs incremental causal graphs for the subsequent time intervals to capture the evolving state of the system behavior across time during an attack event. To mitigate catastrophic forgetting and prioritize relevant historical knowledge, a prior-knowledge-pruned replay buffer is implemented, focusing on subgraphs associated with past attacks and impact points. Causal Edge Reinforcement (CER) is introduced to dynamically adjust edge weights, emphasizing frequently observed attack patterns and mitigating spurious relationships arising from data distribution shifts while maintaining the spectral properties and acyclicity of the causal graphs. 
Finally, in the deep graph convolutional network-based graph classification stage, a Deep Graph Convolutional Network (DGCNN) is employed to distinguish between normal and attack system states, providing a robust and adaptive solution for online anomaly detection in critical infrastructure environments. Figure 2 provides a high-level architecture overview of the three modules in INCADET.

\begin{figure*}
  \centering
  \includegraphics[height=4.5cm, width=\textwidth]{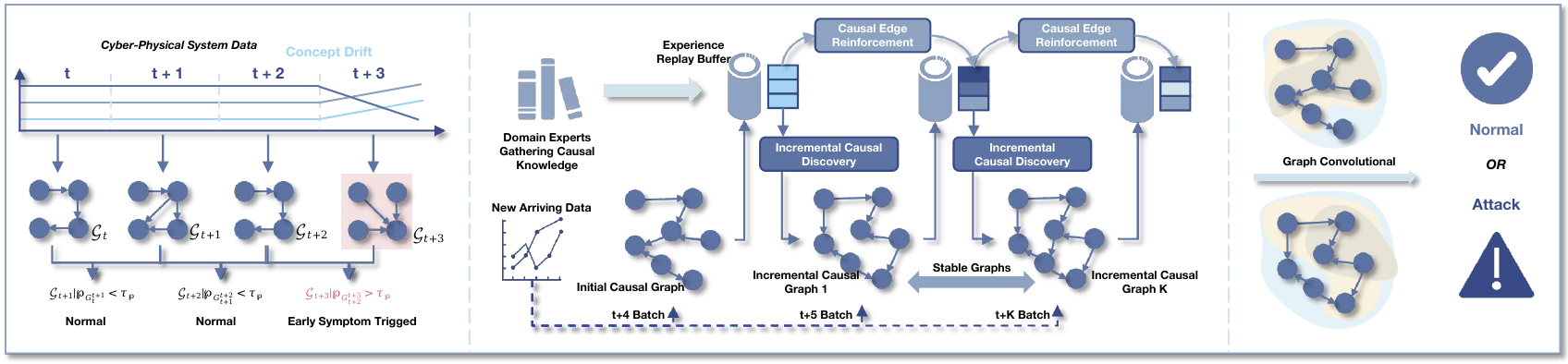}
  \caption{\textbf{Overview of the proposed framework INCADET. INCADET detects early attack symptoms, constructs a causal graph from streaming data, and incrementally updates it using the Experience Replay Buffer and Causal Edge Reinforcement until convergence. Leveraging ground truth causal graphs representing system behavior, a Deep Convolution Graph Neural Network effectively detects and classifies the system status in near real-time.}}
\end{figure*}

\subsection{Early Symptom Trigger Detection}
\textbf{Why does Early Symptom Detection matter?}
Continuously performing incremental causal graph learning throughout the operation of the cyber-physical system will cause unnecessary overhead and reduce system performance.
An early symptom detection module is essential for identifying the optimal trigger point to initiate incremental learning before a system state transitions into an anomalous condition. Without a robust detection mechanism, delayed or premature triggers could either miss critical anomalies or introduce unnecessary computational overhead. Specifically, we quickly compute the static causal graph in each time step and quantize the causal graph through edge weight distribution. We compare the gap between the causal graphs of two consecutive time steps and trigger the incremental causal graph learning mechanism when the gap exceeds a specified threshold.

In this module, our goal is to develop an early symptom detection mechanism that effectively identifies the optimal trigger point for initiating the incremental learning framework. 
We begin by training our model using feature-engineered multivariate time-series sensor data, segmenting the data stream into fixed time windows. The $k^{th}$ time window is given by $\mathbf{X}_k = \{x^i_{t-k+1}, x^i_{t-k+2},...,x^i_{t-1}, x^i_t\}$. 
The initial system state is presumed to be in a 'Normal' operational condition. 
To detect deviations, we construct a static causal graph $G_{t}$ for the time-segment $t$ and subsequently $G_{t+1}$ for the time-segment $t+1$. 
We extract the causal edge weight distribution of the two graphs as $\omega_(G_{t})$ and $\omega_(G_{t+1})$ respectively, The edge weight distribution quantifies the relative strength of causal relationships between system variables within a given time window. Here, edge weights represent the strength of causal relationships between variables in the system, with higher weights indicating stronger influences. This distribution is computed by discretizing the edge weights mapped into $k$ bins and normalizing the probabilistic distribution of edge weights per bin.
To quantify structural changes between consecutive graphs with their respective causal edge weight distribution, we define a graph comparison metric $\wp_{G}$ using the Jensen-Shannon Divergence \cite{1365067} of the two causal edge weight distributions. A significant divergence indicates a potential shift in system behavior, signaling an early symptom trigger.
\begin{equation}
    \wp_{G^{t+1}_t} = 1 - \mathcal{JS}((\omega(G_{t}) || (\omega(G_{t+1}))
\end{equation}
$\wp_{G^{t+1}_t}$ value ranges from $[0 \sim 1 ]$, where a value closer to $1$ indicates higher similarity between the graphs, while a value closer to $0$ signifies greater divergence.
Based on above equation, we compute the $\wp_{G^{t+1}_t}$ for each subsequent graph. We set a threshold value $\tau_{\wp}$ and determine the trigger point $\mathcal{T}$ which indicates that the static causal graph at the time step is significantly divergent from a system during "Normal" behavior. 
Our analysis showed that our module effectively identifies the critical precursors to attack events, leveraging causal structural and edge weight changes to substantially improve early threat detection capabilities.
\begin{equation}
    \mathcal{T} = G_{t_k} | \wp_{G^{t_k}_{t_{k-1}}} > \tau_{\wp}  
\end{equation}

Upon determining the early symptom trigger graph, the incremental causal learning framework is initiated to learn the continual causal graph representing anomalous system behavior. 
Once the stopping criteria of incremental learning is satisfied, the module transitions back to determine the next trigger point of an attack event.

\subsection{Incremental Causal Graph Construction}
\textbf{Why do we need Incremental Causal Graphs?}
Once an early symptom trigger is detected, the next challenge is efficiently capturing the evolving causal structure of the system in response to anomalies. Traditional causal discovery methods struggle in streaming environments due to catastrophic forgetting, memory overhead, and sensitivity to distribution shifts. To address these challenges, we propose an incremental causal graph learning module that dynamically updates causal representations while preserving high-value historical knowledge.
Specifically, this module constructs an incremental causal graph that captures real-time system changes while retaining critical causal relationships from past anomalous events. Upon detecting a trigger point, we leverage Bayesian graph learning, experience replay, and causal edge reinforcement to maintain a robust and interpretable causal structure.

Our goal is to efficiently represent causal knowledge from the dynamic data stream based on system status. Specifically, this module builds an incremental causal graph that characterizes the anomalous or “Attack” system status. The graph not only reflects causal relationships in newly observed data but also inherits high-value historical knowledge, preventing catastrophic forgetting. Once the trigger time segment is identified by the early symptom detection module, the incremental causal graph module is activated to adaptively update the system’s causal understanding.

 Given a time window of data $\mathbf{X}_k$, we construct a causal graph representing anomalous system behavior from $\mathbf{X}_k$ and the time-lagged data $T$ using DYNOTEARS \cite{pamfil2020dynotears} Bayesian Graph Learning algorithm. The Structural Equation Model (SEM) is formulated as follows.
\begin{equation}
    X_k = X_kA_{X_k} + T_kA_{T_k} + Z_k
\end{equation}
where,
the errored variables are regularized by the vector $Z$.

DYNOTEARS considers this as an optimization problem to estimate the weighted adjacency matrices $A_X$ and $A_T$ and is formulated as a function $f(A_X, A_T)$:
\begin{equation}
    \text{min}_{A_X,A_T} f(A_X,A_T) \text{ s.t. $A_X$ is acyclic}
\end{equation}
\begin{equation}
    \text{with } f(A_X,A_T) = \ell(A_X,A_T) + \lambda_{A_X}||A_X||_1 + \lambda_{A_T}||A_T||_1
\end{equation}
 
To enforce sparsity to the adjacency matrices, $\ell_1$ penalties have been added to the above objective function.

From the incoming stream of data at time window $k$, the initial causal graph is given by $G^k(\mathcal{V}_{attack}, \mathcal{E}_{attack})$ denoted as $G^k_{A}$.
To incrementally update causal knowledge, we propose a methodology using Experience Replay and Edge Reinforcement to incorporate crucial knowledge from past anomalous events. Given $\mathbf{X}_k$ and the corresponding causal graph $G^k_A$, we aim to extract the important causal knowledge $\hat{U}_k$ representing an attack event during the time window $k$ and incrementally update the causal graph $G^{k+1}_A$ constructed at the future time window $k+1$.
\begin{equation}
    \hat{U}_k = \delta(G^k_A)
\end{equation}

\textbf{Prior-Knowledge Pruned Replay Buffer.} In incremental causal learning, preserving past information is essential for updating the current graph construction, allowing the model to retain historical causal knowledge and adapt to evolving system behavior. However, indiscriminately storing past data can lead to significant memory and computational overhead, especially when retaining less informative knowledge. To address this challenge, we propose a prior-knowledge-driven replay buffer that selectively stores high-value causal information to ensure efficient and scalable updates. To tackle this challenge, We use prior attack knowledge (attack/impact points) to extract a subgraph $\hat{G}^k_A$ from $G^k_A$. We denote the replay buffer as $\mathbf{E}_k = \{(A_1, I_1),(A_1, I_2), (A_2, I_2), (A_2, I_3),...\}$ where $A_i$ represents the attack points and $I_i$ represents the impact points obtained by the directed edges of the subgraph $\hat{G}^k_A$. Given the prior knowledge of the impacted nodes $\mathcal{D}$ and the constructed causal graph $G^k_A$, the subgraph $\hat{G}^k_A$ is extracted as below.
\begin{equation}
    \hat{G}^k_A = G^k_A \{(\hat{u},\hat{v}) \, | \, \hat{u} \in \mathcal{D},\, \hat{v} \in \mathcal{D},\, and \, (\hat{u},\hat{v}) \in \mathcal{E}^k_{attack}\} 
\end{equation}
From the extracted sub-graph, we store the directed edge set $\mathcal{\hat{E}}^k_{attack}$ and the edge weights $\mathcal{W} = \{w_1, w_2, w_3, ...\}$ in the replay buffer $\mathbf{E}_k$ such that $\mathbf{E}_k = \{(\hat{u}_1,\hat{v}_1, w_1), \, (\hat{u}_2,\hat{v}_2, w_2), ....\}$. 

\textbf{Causal Edge Reinforcement (CER).} In real-time scenarios, evolving data distributions and attack events induce structural shifts in causal graph construction. A naïve approach of adding causal edges from the replay buffer without adaptation can distort critical knowledge, leading to suboptimal data representations. Furthermore, frequent attack events in early time windows may later be identified as spurious correlations due to changing system behavior. To tackle this, we propose Causal Edge Reinforcement, a mechanism that incrementally assigns higher weights to causal edges corresponding to frequent attack events. This approach ensures that critical attacks are prioritized over rare occurrences, effectively mitigating spurious relationships introduced by distribution shifts in time-series data.

To preserve the spectral properties of the causal graph, we employ the Weighted Laplacian Transformation for incremental updates using past knowledge. At each time window, we extract $\mathbf{E}_k$, the set of impacted edges from past events. If these causal edges reappear in the future graph $G^{k+1}_A$, their edge weights are reinforced by a constant $\omega$. Otherwise, we simply add those edges to the graph $G^{k+1}_A$ to avoid catastrophic forgetting of that knowledge over time.

For a causal graph $G^k_A$, its Laplacian representation $\mathcal{L}(G^k_A)$ is given by
\begin{equation}
    \mathcal{L}(G^k_A) = D(G^k_A) - A(G^k_A)
\end{equation}
where,
$D(G^k_A)$ is the Degree matrix of the graph and $A(G^k_A)$ is the Adjacency matrix of Graph $G^k_A$. 

For the edges $\mathcal{E}_(i,j)$ extracted from $\mathbf{E}_k$, we apply the causal edge reinforcement as follows. 

\begin{equation}
    \widehat{A}(G^{k+1}) = 
    \begin{cases}
    w^k_{i,j} \times  \omega, & \forall \mathcal{E}_{i,j} \in \mathcal{E}_k, \, w_{i,j} \in \mathcal{W} \\
    w^{k+1}_{i,j}, & \text{otherwise}
    \end{cases}
\end{equation}

With the causal edge reinforcement, for the causal graph $G^{k+1}_A$ at time window $k+1$, its Laplacian representation $\mathcal{L}(G^{k+1}_A)$ is given by
    \begin{equation}
         \mathcal{L}(G^{k+1}_A) = D(G^{k+1}_A) - \widehat{A}(G^{k+1}_A)
    \end{equation}

As the scale grows, certain node degrees impacted by the attack events might outgrow to an extent where it induces overwhelming bias in the graph convolution process. So we utilize the Symmetric Laplacian Normalization technique to mitigate the bias and enhance generalization. It is given by

\begin{equation}
    \mathcal{L}_{norm}(G^{k+1}_A) = D(G^{k+1}_A)^{(-1/2)} * \mathcal{L}(G^{k+1}_A) * D(G^{k+1}_A)^{(-1/2)}
\end{equation}

Since the constructed causal graphs are Directed Acyclic Graphs, adding incremental edges might disturb the essential acyclicity constraints. In order to tackle this, we introduce Protected Nodes-based cycle removal where we remove the cycles from a graph while preserving the edges from the nodes representing attack events from the domain knowledge replay buffer $E_K$

\textbf{Stopping Criteria.} The constructed incremental graph gradually converges over time such that the outdegree distributions of the subgraph $\hat{G}^k_A$ will be closer to that of the subgraph $\hat{G}^{k+1}_A$ obtained at the time window k+1 given the domain knowledge $\mathcal{D}$. We utilize the JS Divergence over the edge degree distributions to impose the stopping criteria $\mathcal{S}$ with a threshold $\tau_{\mathcal{\hat{E}}}$.
\begin{equation}
    \mathcal{S} = 1-\mathcal{JS}(\mathcal{\hat{E}(G^k_A)}||\mathcal{\hat{E}(G^{k+1}_A)})
    \mathcal{S}<\tau_{\mathcal{\hat{E}}}
\end{equation}

For constructing a causal graph representing the "Normal" status, we follow a similar approach as above but the difference here is the causal sub-graph extracted based on the domain knowledge of attack and impact points is significantly different from that of the one representing the "Attack" status. We utilize this evident difference in causal structure to efficiently construct causal representations of the system status.

\subsection{Deep Graph Convolutional Network-based Graph Classification}
While incremental causal graph learning provides a structured representation of system evolution, an additional challenge is efficiently classifying system states based on these evolving structures. Traditional point-based anomaly detection models struggle to capture complex dependencies in dynamic environments. In contrast, graph-based classification frameworks leverage topological patterns, improving robustness to noise and enhancing interpretability.

To achieve this, we employ a Deep Convolutional Graph Neural Network (DGCNN) \cite{10.5555/3504035.3504579} for classifying system states using incremental causal graphs. The DGCNN model processes Laplacian representations, attack labels, and edge indices, extracting hierarchical graph features to distinguish between normal and attack states. The following section details the mathematical formulation and implementation of our graph classification module.

\begin{equation}
    H^{(i+1)} = \sigma(L_{agg} * H^{(i)} * W^{(i)})    
\end{equation}
where,
$H^{(i)}$ is the $i^{th}$ hidden layer in the network, $\sigma$ refers to the non-linearity function ReLU, and $W$ is the corresponding weight matrix.

We use the Binary Cross Entropy (BCE) Loss as the loss function of our graph classification module. It is given by
\begin{equation}
    \mathcal{L} = 1/N \sum_{i=1}^N -(y_i * log(p_i) + (1-y_i) * log(1-p_i))
\end{equation}
where,
$y_i$ is the actual label, and $p_i$ is the predicted probability of the sample $i$ in the positive class given by the sigmoid function for binary graph classification.

For the test data split by fixed time-windows k, we construct a causal graph using DYNOTEARS \cite{pamfil2020dynotears} representing the causal structure of the data stream in the given time segment, and the trained DGCNN model is applied to classify the test graph into its corresponding attack status.

\subsection{Implementation Details}
\textbf{Assumptions: } An implicit assumption in real-time anomaly detection research task is that "Attack" events will occur far less frequently than that of "Normal" system status. This data imbalance perturbs the efficiency of many existing statistical-based approaches in real-world scenarios. We also make an assumption that a reliable industrial cyber-physical system is less likely to have all the nodes and sensors susceptible to attack events. In constructing incremental graphs, we assume that the causal structure representing the "Normal" system status doesn't exhibit drastic structural changes over time.  

\textbf{Implementation: } Firstly, we model the incoming time-series sensing data into fixed batches of data split by time window $k$ of $15$ minutes. The framework is trained on the historical time series data and the test data is the future data adhering to the data model and learning model assumptions. From the datasets, we also collect prior knowledge on the various cyberattack events including temporal attack patterns, attack and impact nodes, and attack purposes.

To construct the causal graphs, we utilized the CausalNex package to implement the DYNOTEARS \cite{pamfil2020dynotears} with a time-lag parameter of $4$ for SWaT,$3$ for WADI, $4$ for TE, and $1$ for SMD datasets respectively. This hyperparameter setting is proved in experiments to be optimal in capturing the delayed and cascading impacts of cyberattack events. For the early symptom trigger detection, we used a $\tau_{\wp}$ value of $\sim0.9$. Inspired from \cite{10.1145/3627673.3680096}, we utilized weighted Laplacian transformation to implement Causal Edge Reinforcement to fuse the domain knowledge subgraph into the future incremental causal graph. The stopping criteria for the incremental graph construction is given by $\tau_{\mathcal{\hat{E}}}$ = 0.1. In the DGCNN module, we implemented 3 Convolution Layers with a Rectified Linear Unit (RELU) activation function with a dropout layer. We used Adam Optimizer and a learning rate of $0.01$ to optimize the performance of our framework. The binary classification output of this module is given by a sigmoid function representing "Attack" and "Normal" statuses respectively.

For the experiments, the graph-based frameworks were implemented using the original code from the published work with default hyperparameter settings. The non-graph-based models were experimented using PyOD outlier detection \cite{zhao2019pyod} implementations.

\section{Experiments}
This section presents a comprehensive experimental evaluation of INCADET, designed to address the following research questions.
\begin{enumerate}\renewcommand{\labelenumi}{\textbf{RQ\arabic{enumi}.}}
    \item Does INCADET perform on par with the other baseline algorithms for real-time temporal anomaly detection?
    \item Why and how is causal learning better than the traditional Deep Learning approaches to perform anomaly detection?
    \item To what extent does our framework's performance depend on the model hyperparameters?
    \item How reliable is INCADET when it comes to addressing scalability challenges?
    \item How efficient is our Incremental Learning framework over the Offline Causal Models in cyberattack detection?   
\end{enumerate}

\subsection{Experimental Setup}
\subsubsection{Datasets. } Extensive experiments were conducted using our framework on three real-world testbed datasets: \textbf{1) SWaT} \cite{7469060}: A 11-day water treatment plant testbed data collected from 51 interconnected sensors. The dataset contains 16 fault events over the period which had 7 days of "Normal" system status and 4 days of "Attack" system status. \textbf{2) WADI} \cite{10.1145/3055366.3055375}: Water Distribution testbed dataset collected over 16 days from 123 actuators/sensors. 15 attack events were recorded in the last 2 days of the collection period with "Attack" system status. \textbf{3) Tennessee Eastman (TE)} \cite{4519-z502-19}: A chemical process simulation dataset with 52 sensors and 20 anomalies. Training spans 25 hours, testing 48 hours, with measurements every three minutes. Table 1 shows dataset statistics, train-test split, and causal graph node count. \textbf{4)Server Machine Dataset (SMD) \cite{10.1145/3292500.3330672}} is a server monitoring dataset over 5 weeks monitoring 28 server machines from 38 sensors. It is one of the largest public datasets for anomaly detection in multivariate time-series data.

\begin{table}[h!]
\renewcommand\arraystretch{1.1}
	\centering
	\caption{DATASET STATISTICS.}         
	\vspace{-0.1cm}
        \scalebox{0.70}{
	\begin{tabular}{c|c|c|c|c|c}
		\hline
		  Dataset & Status & \# Features & \# Normal Data & \# Attack Data & Normal to Attack Ratio        \\ \hline
            \textbf{SWaT} & Normal & 51 & 495000 & 0 & 0        \\
                 &Attack & 51 & 395298 & 54621 & 7:1        \\
            \hline
            \textbf{WADI} & Normal & 123 & 1048571 & 0 & 0        \\
                 &Attack & 123 & 162824 & 9977 & 16:1        \\
            \hline
            \textbf{TE} & Normal & 52 & 450000 & 0 & 0        \\
                 &Attack & 52 & 222500 & 22800 & 10:1        \\
            \hline
            \textbf{SMD} & Normal & 38 & 708405 & 0 & 0        \\
                 &Attack & 38 & 678950 & 29470 & 25:1        \\
            \hline
	\end{tabular}
        }
	\label{tab:dataset statistics}
\end{table}

\subsubsection{Evaluation Metrics}
The performance of the proposed framework is evaluated using the following widely-used metrics.

\textbf{Point-adjusted F1-Score \cite{febrinanto2023entropy}: } In multivariate time-series data, attack events often form continuous segments, making pointwise anomaly detection less effective \cite{9525836}. Our causal framework also requires a time-interval segment to capture the underlying causal structure. To address this, we use the point-adjusted F1-Score, where a segment is labeled anomalous if any point within it is an anomaly, and the F1-score is computed based on the framework’s performance on the entire segment.
\textbf{ROC-AUC: }AUC, the area under the Receiver Operating Characteristic (ROC) curve, provides a measure of a model's ability to distinguish positive from negative samples.
\textbf{PRC-AUC: }PRC-AUC focuses on the performance of the model on anomalous class. This is particularly useful when the dataset is imbalanced like in anomaly detection.
\textbf{Mean Absolute Error (MAE):}Proportion of normal data instances that the model predicts incorrectly.
\textbf{Missed Alarm Rate (MAR): }Proportion of actual anomalies classified as normal by the model. It is a compliment to the Sensitivity of the model.

\subsubsection{Baselines}
We evaluate the performance of our INCADET framework with the following baseline algorithms.
\textbf{USAD} \cite{10.1145/3394486.3403392} – Uses adversarial training with two autoencoders to detect anomalies via reconstruction error.
\textbf{DAGMM} \cite{DBLP:conf/iclr/ZongSMCLCC18} – A Deep Autoencoder Gaussian Mixture Model leveraging expectation-maximization for anomaly detection.
\textbf{MTAD-GAT} \cite{Zhao_2020} – A self-supervised Graph Attention Network optimizing forecasting and reconstruction models.
\textbf{Isolation Forest} \cite{xu2017improved} – A supervised ensemble method using sub-sampling to explicitly isolate anomalies.
\textbf{LODA} \cite{10.1007/s10994-015-5521-0} – A lightweight ensemble model of weak unsupervised outlier detectors.
\textbf{ABOD} \cite{10.1145/1401890.1401946} – An Angle-Based Outlier Detection method using cosine angles to identify deviations.
\textbf{Deep SVDD} \cite{zhang2021anomaly} – A deep learning-based Support Vector Data Description method using a hypersphere for anomaly detection.
\textbf{SMV-CGAD} \cite{10.1145/3627673.3680096} – A Sparsified Multi-View Causal Graph-based framework modeling anomalous events with temporal dependencies.
We use the default hyperparameters for all baselines as reported in their original works.

\subsubsection{Experimental Settings}
All the experiments and evaluations were carried out on an Intel i9-12900HK 2.50GHz x64-based processor with 32.0 GB RAM and GeForce RTX 3050 Ti GPUs. The implementations used Python 3.10.4 and PyTorch 1.8.2 frameworks.

\subsection{Performance Evaluation}

\subsubsection{Overall Performance}
Extensive experiments show that INCADET performs competitively against various time-series anomaly detection frameworks, including supervised, unsupervised, graph-based, and offline causal graph-based models. Addressing \textbf{RQ1}, Table 2 demonstrates that INCADET outperforms baseline algorithms in Point-Adjusted F1-score and PRC-AUC across all datasets. It also leads in ROC-AUC except on the SMD dataset, where Isolation Forest performs better due to its lower dimensionality and simpler noise conditions, where ensemble methods excel. This highlights that while INCADET excels in complex, high-dimensional scenarios, simpler models can remain effective in less demanding environments. Causal graph-based frameworks, SMV-CGAD and INCADET, improve ROC-AUC and PRC-AUC by $\sim2\%,\, \sim10.5\%,\, \sim5.5\%,$ and $\sim4\%$ on SWAT, WADI, TE, and SMD datasets, respectively. These consistent improvements across diverse datasets underscore the inherent robustness of causal models in capturing complex dependencies and handling varying data characteristics. The superior performance of causal frameworks is attributed to their ability to handle distribution shifts and spurious associations, along with incorporating prior knowledge. The baseline algorithms struggle with high data imbalance and distribution shifts, resulting in unstable performance. This instability reveals the limitations of traditional methods in adapting to real-world, non-stationary data. The poor performance of the baseline algorithms on the WADI dataset is likely due to its high dimensionality.  This illustrates the "curse of dimensionality," which particularly affects traditional approaches when applied to imbalanced anomaly detection datasets. As observed, despite USAD's strong performance on other datasets, its effectiveness is diminished on WADI due to the challenges posed by high dimensionality and noisy data. The geometric measurement-based approaches like ABOD and Deep SVDD showed poor performance when the scale of the features increased. This suggests that these methods are sensitive to feature space expansion, potentially due to increased sparsity and reduced density in high-dimensional spaces. DAGMM's strict Gaussian assumption and low-dimensional latent space often lead to information loss, hindering its performance in real-world scenarios. This points to the importance of model flexibility and the ability to capture non-Gaussian data distributions in practical anomaly detection. MTAD-GAT, a state-of-the-art graph attention framework, has shown strong performance against other baselines. However, the framework is a complex "black-box" model and the interpretability of the decisions becomes difficult in real-world cases. The trade-off between performance and interpretability is a critical consideration in deploying anomaly detection systems, particularly in critical infrastructure where explainability is paramount. A key advantage of our causal framework, in contrast to traditional black-box models, is its use of incremental causal relationships to represent evolving events \cite{gu2020causality}.  By considering these causal links as a deciding factor in attack detection, the model provides transparency into its reasoning. This directly addresses \textbf{RQ2}, showcasing the greater interpretability of causal models compared to traditional approaches. The explicit representation of causal relationships in INCADET not only enhances interpretability but also facilitates actionable insights, enabling operators to understand the root causes of anomalies and implement targeted mitigation strategies.

\begin{table*}
\renewcommand\arraystretch{1.2}
\centering
\caption{PERFORMANCE ANALYSIS}
\label{tab:performance_analysis}
\small{
\begin{tabularx}{\textwidth}{c|X|X|X|X|X|X|X|X|X|X}
\hline
\multirow{2}{*}{Methods} & \multicolumn{5}{c|}{\textbf{SWAT}} & \multicolumn{5}{c}{\textbf{WADI}} \\
\cline{2-11}
                         & $F1_{PA}$ & ROC-AUC & PRC-AUC & MAR & MAE & $F1_{PA}$ & ROC-AUC & PRC-AUC & MAR & MAE \\
\hline
USAD                     & 0.8155 & 0.8270 & 0.7843 & \textbf{0.2612} & 0.1190 & 0.4326 & 0.7176 & 0.4860 & 0.4228 & 0.4486\\
DAGMM                    & 0.6129 & 0.7881 & 0.4997 & 0.5450 & 0.5241 & 0.1491 & 0.6038 & 0.1798 & 0.8042 & 0.7612\\
MTAD-GAT                 & 0.7003 & 0.7249 & 0.7320 & 0.3761 & 0.3245 & 0.3877 & 0.7397 & 0.3137 & 0.6366 & 0.6899 \\
LODA                     & 0.6942 & 0.5572 & 0.7154 & 0.3818 & 0.3748 & 0.2451 & 0.5516 & 0.1920 & 0.5602 & 0.3702 \\
ABOD                     & 0.6094 & 0.5000 & 0.6321 & 0.0801 & 0.9390 & 0.1126 & 0.5000 & 0.1753 & 0.9761 & \textbf{0.1566} \\
Deep-SVDD                & 0.1785 & 0.6632 & 0.2441 & 0.4410 & 0.4823 & 0.1643 & 0.4288 & 0.2121 & 0.4745 & 0.5790 \\
Isolation Forest         & 0.7512 & 0.8026 & 0.8122 & 0.3788 & 0.2214 & 0.6034 & 0.6407 & 0.5850 & 0.6623 & 0.5130 \\
SMV-CGAD                 & 0.7532 & 0.8211 & 0.7355 & 0.3244 & \textbf{0.0900} & 0.6679 & 0.7831 & 0.7077 & 0.3100 & 0.2446 \\
\textbf{INCADET}         & \textbf{0.8179} & \textbf{0.8300} & \textbf{0.8388} & 0.2711 & 0.1131 & \textbf{0.6733} & \textbf{0.8045} & \textbf{0.7104} & \textbf{0.2898} & 0.1844 \\
\hline
\end{tabularx}
}
\end{table*}
\begin{table*}
\renewcommand\arraystretch{1.2}
\centering
\label{tab:performance_analysis}
\small{
\begin{tabularx}{\textwidth}{c|X|X|X|X|X|X|X|X|X|X}
\hline
\multirow{2}{*}{Methods} & \multicolumn{5}{c|}{\textbf{TE}} & \multicolumn{5}{c}{\textbf{SMD}} \\
\cline{2-11}
                         & $F1_{PA}$ & ROC-AUC & PRC-AUC & MAR & MAE & $F1_{PA}$ & ROC-AUC & PRC-AUC & MAR & MAE \\
\hline
USAD                     & 0.7258 & 0.7313 & 0.6851 & 0.3226 & 0.3617 & 0.8151 & 0.8334 & 0.8200 & 0.2876 & 0.2219 \\
DAGMM                    & 0.6312 & 0.7280 & 0.5680 & 0.1460 & 0.4478 & 0.6410 & 0.7185 & 0.5485 & 0.4587 & 0.4903\\
MTAD-GAT                 & 0.7592 & 0.8010 & 0.6899 & 0.3228 & 0.3579 & 0.8067 & 0.6891 & 0.7881 & 0.3914 & 0.2326 \\
LODA                     & 0.4228 & 0.5007 & 0.5102 & 0.6212 & 0.5592 & 0.2656 & 0.5030 & 0.2866 & 0.7214 & 0.5641 \\
ABOD                     & 0.1827 & 0.4730 & 0.2560 & 0.7684 & 0.2572 & 0.6244 & 0.5027 & 0.5009 & 0.5186 & 0.4247 \\
Deep-SVDD                & 0.4801 & 0.5679 & 0.4680 & 0.5620 & 0.4801 & 0.3911 & 0.5448 & 0.4794 & 0.5324 & 0.2558 \\
Isolation Forest         & 0.7421 & 0.8102 & 0.7684 & 0.2814 & 0.3197 & 0.8501 & \textbf{0.8734} & 0.8681 & 0.3229 & 0.2406 \\
SMV-CGAD                 & 0.7389 & 0.8002 & 0.7556 & 0.2245 & 0.2765 & 0.8395 & 0.8030 & 0.8112 & 0.2166 & \textbf{0.1909} \\
\textbf{INCADET}         & \textbf{0.7762} & \textbf{0.8612} & \textbf{0.8018} & \textbf{0.2100} & \textbf{0.2398} & \textbf{0.8788} & 0.8622 & \textbf{0.9026} & \textbf{0.1996} & 0.1934 \\
\hline
\end{tabularx}
}
\end{table*}

\subsubsection{Ablation Study}
In this section, we present the experimental results assessing the main components and their effective contribution to our framework. To do so, we conducted an ablation study with the following settings.
\begin{itemize}
    \item \textbf{W/O Replay Buffer}: We exclude the replay buffer that stores the causal edges of cyberattack events from the past. 
    \item \textbf{W/O Causal Edge Reinforcement (CER)}: We exclude the causal edge reinforcement of the causal relationships that occur frequently over time.
    \item \textbf{W/O DYNOTEARS}: We replace temporal causal learning algorithm DYNOTEARS with DAGs with NOTEARS \cite{zheng2018dags}.
\end{itemize}
The experimental results of the study are shown in Table 3. The summary of the findings is as follows.
\begin{itemize}
    \item The replay buffer is critical in preventing catastrophic forgetting within the incremental framework. In the absence of the buffer, performance deteriorates significantly. The results highlight the importance of the replay buffer in effectively preserving historical causal knowledge.
    \item Excluding causal edge reinforcement leads to a modest performance decline. Reinforcing edge weights helps preserve causal relationships in non-stationary data, enabling better generalization and reducing sensitivity to noise.
    \item DYNOTEARS effectively captures time-lagged causal relationships in multivariate temporal settings. Replacing it with DAGs using NOTEARS leads to significant performance degradation, as they cannot model temporal dependencies, thereby failing to capture the crucial delayed impacts.
\end{itemize}


\begin{table*}
\renewcommand\arraystretch{1.2}
\centering
\caption{ABLATION STUDY}
\label{tab:ablation_study}
{\small
\begin{tabularx}{\linewidth}{c|X|X|X|X|X|X} 
\hline
\multirow{2}{*}{Methods} & \multicolumn{3}{c|}{\textbf{SWAT}} & \multicolumn{3}{c}{\textbf{WADI}} \\
\cline{2-7}
 & $F1_{PA}$ & ROC & PRC & $F1_{PA}$ & ROC & PRC \\
\hline
INCADET                     & \textbf{0.8179} & \textbf{0.8300} & \textbf{0.8388} & \textbf{0.6733} & \textbf{0.8045} & \textbf{0.7104} \\
w/o Replay Buffer           & 0.4617 & 0.5010 & 0.3997 & 0.2411 & 0.5038 & 0.2798 \\
w/o CER                     & 0.7713 & 0.8008 & 0.7524 & 0.6427 & 0.7014 & 0.7436 \\
w/o DYNOTEARS               & 0.1761 & 0.5102 & 0.1254 & 0.0901 & 0.4750 & 0.1120 \\
\hline
\end{tabularx}
}
\end{table*}
\begin{table*}
\renewcommand\arraystretch{1.2}
\centering
\label{tab:ablation_study}
{\small
\begin{tabularx}{\linewidth}{c|X|X|X|X|X|X} 
\hline
\multirow{2}{*}{Methods} & \multicolumn{3}{c|}{\textbf{TE}} & \multicolumn{3}{c}{\textbf{SMD}} \\
\cline{2-7}
 & $F1_{PA}$ & ROC & PRC & $F1_{PA}$ & ROC & PRC \\
\hline
INCADET                     & \textbf{0.7762} & \textbf{0.8612} & \textbf{0.8018} & \textbf{0.8788} & \textbf{0.8622} & \textbf{0.9026} \\
w/o Replay Buffer           & 0.4119 & 0.4881 & 0.3926 & 0.2491 & 0.5002 & 0.3745 \\
w/o CER                     & 0.7433 & 0.8202 & 0.7910 & 0.8467 & 0.8021 & 0.8314 \\
w/o DYNOTEARS               & 0.2242 & 0.5001 & 0.2476 & 0.1402 & 0.4591 & 0.1118 \\
\hline
\end{tabularx}
}
\end{table*}

\subsubsection{Sensitivity to Hyperparameters}
In this section, we present an experiment analysis of the hyperparameters and its influence on the efficiency of our framework (\textbf{RQ3}). We employ a random search technique to optimize hyperparameters for our framework's performance. In the causal graph learning phase, we use the DYNOTEARS algorithm, where the time-lag parameter $\tau$ is crucial for capturing delayed dependencies in the data. As seen in Fig. 3, a small $\tau$ misses delayed attack impacts, while a large $\tau$ risks overfitting, and the optimal $\tau$ is determined based on data distribution and attack-event timelines. In the incremental learning phase, we apply the weighted Laplacian transformation for causal edge reinforcement, with the weight parameter $\omega$ controlling the importance of causal edges in attack knowledge. We set $\omega = 2$ across datasets to prevent overfitting. In the DGCNN module, the learning rate ($\alpha$) and the choice of optimizer are critical, with $\alpha = 0.01$ and the Adam optimizer yielding the best performance, demonstrated by smooth ROC curves.

\subsubsection{Scalability and Required Data Volume Analysis of Incremental Learning}
Fig. 4 illustrates the framework's performance on various datasets across different scales of streaming data, providing an overview of its scalability. The performance significantly improves as the incoming training batches increase, attributed to the integration of Causal Edge Reinforcement and the Replay Buffer, which effectively eliminates spurious edges in the causal graphs. This approach reduces memory consumption by preventing the need for retraining on the entire past dataset, addressing catastrophic forgetting, and enabling efficient performance at larger scales (\textbf{RQ4}).

\begin{figure*}[htbp]
\centering
\subfigure[Performance vs. Timelag on SWaT]{
\includegraphics[width=4.3cm, height=3.5cm]{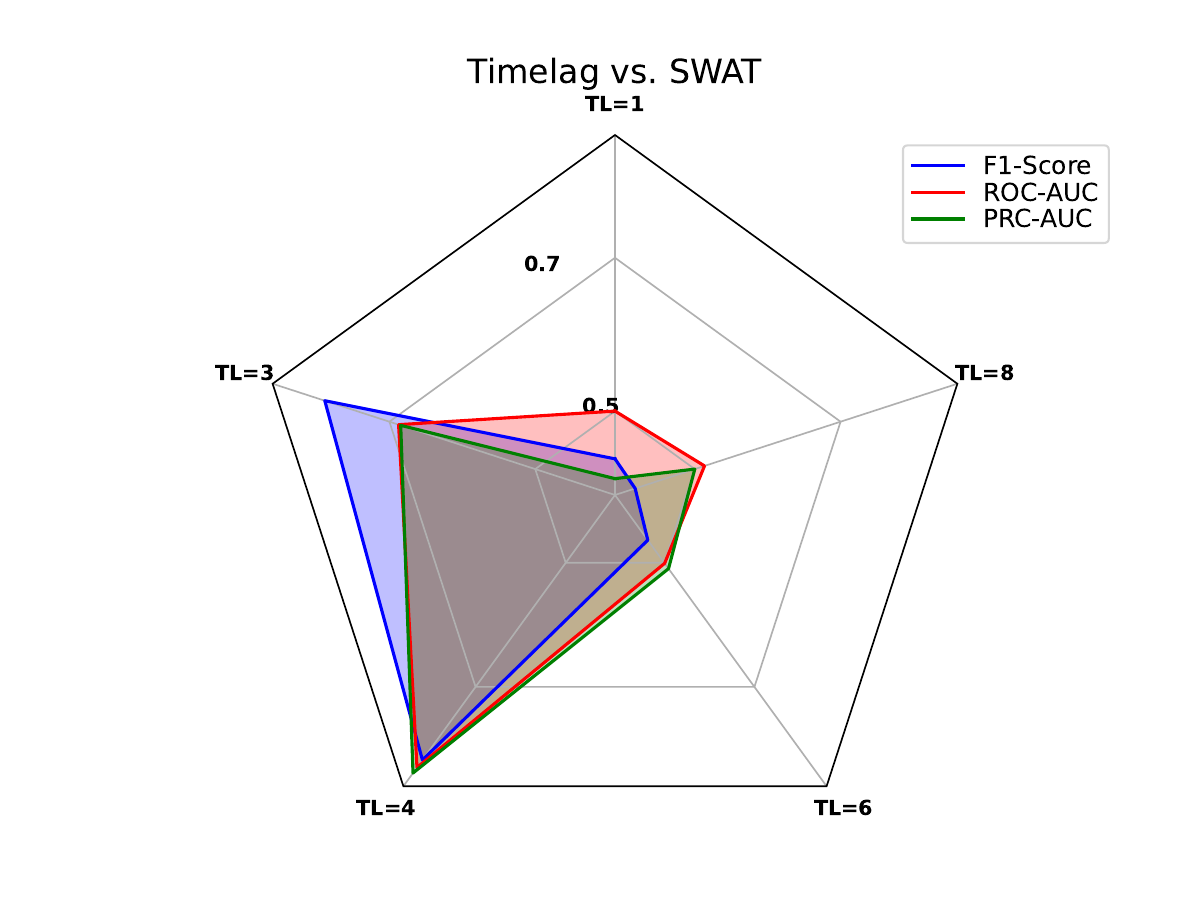}
}
\hspace{-3mm}
\subfigure[Performance vs. Timelag on WADI]{ 
\includegraphics[width=4.3cm, height=3.5cm]{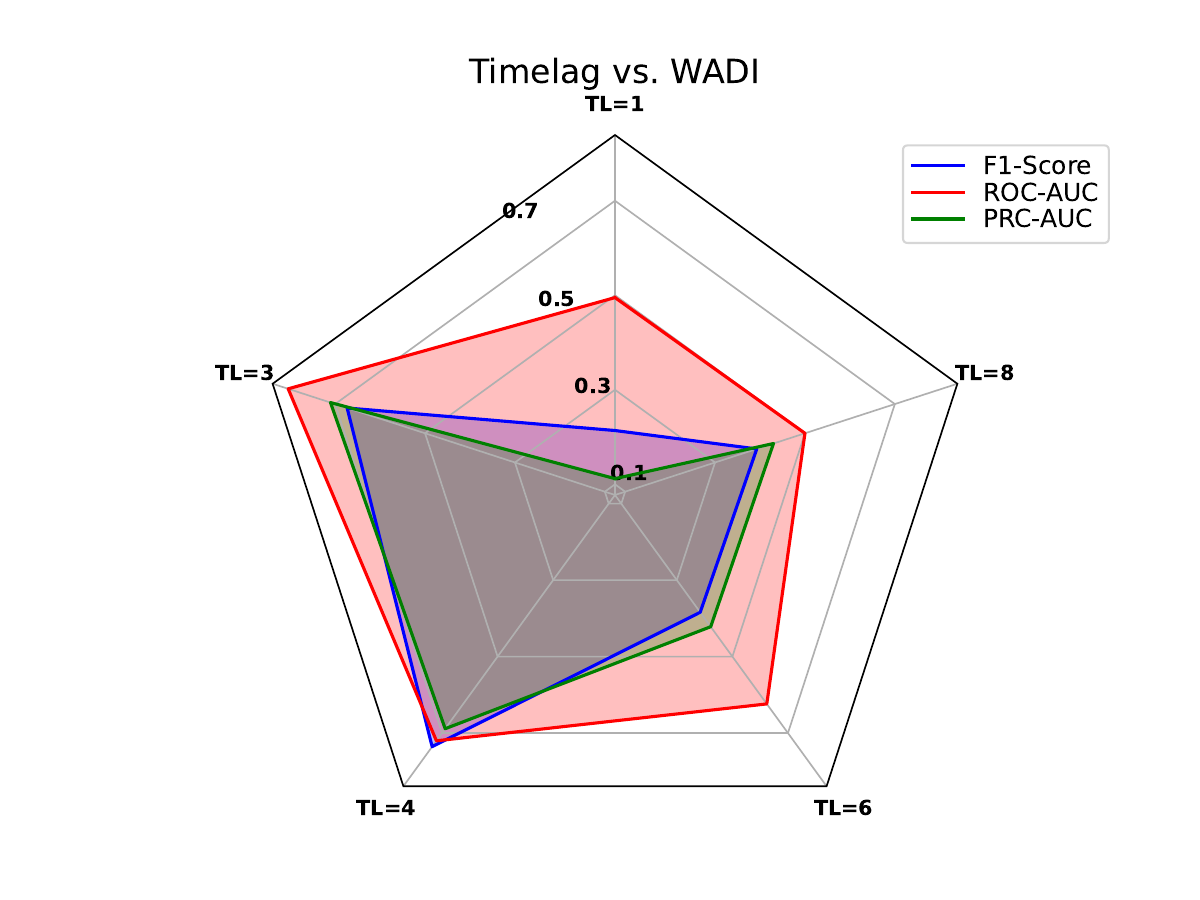}
}
\hspace{-3mm}
\subfigure[Performance vs. Timelag on TE]{ 
\includegraphics[width=4.3cm, height=3.5cm]{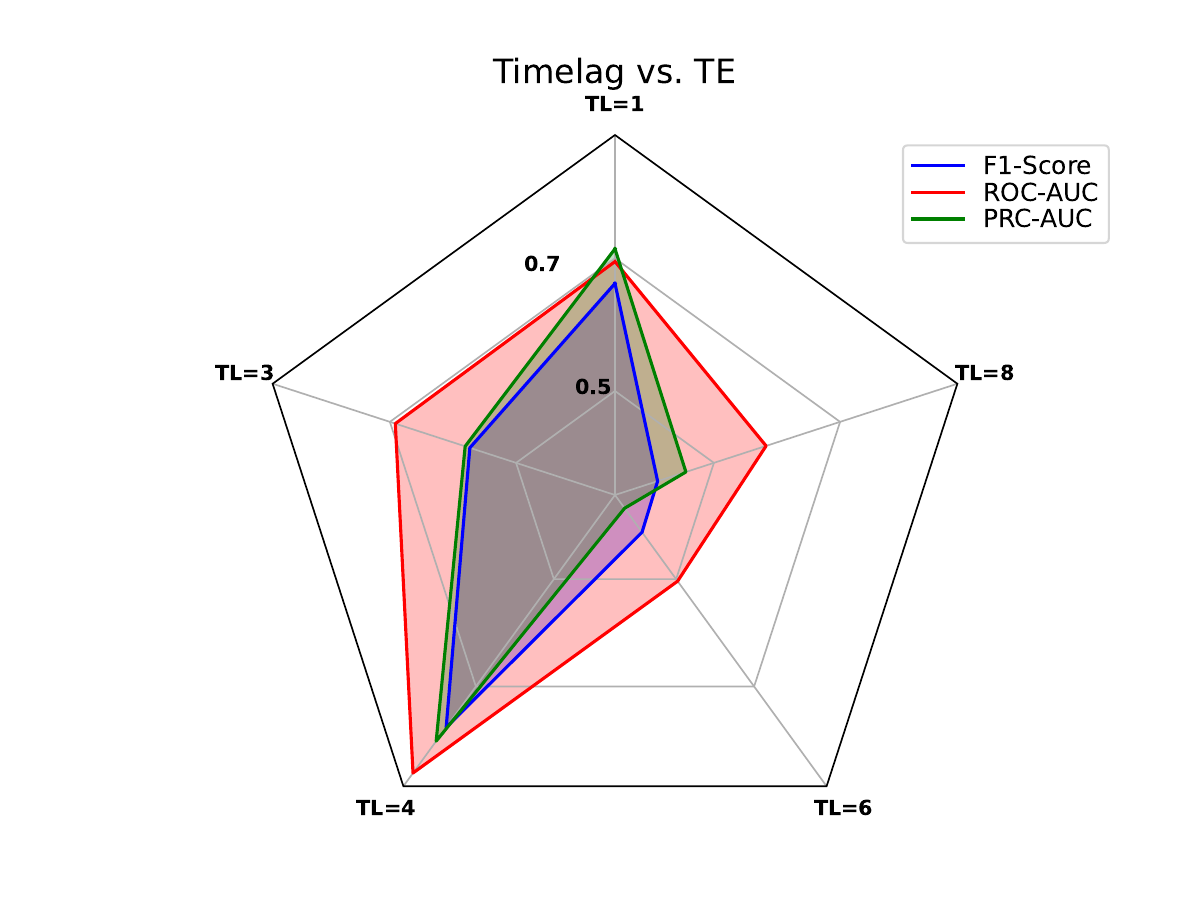}
}
\hspace{-3mm}
\subfigure[Performance vs. Timelag on SMD]{ 
\includegraphics[width=4.3cm, height=3.5cm]{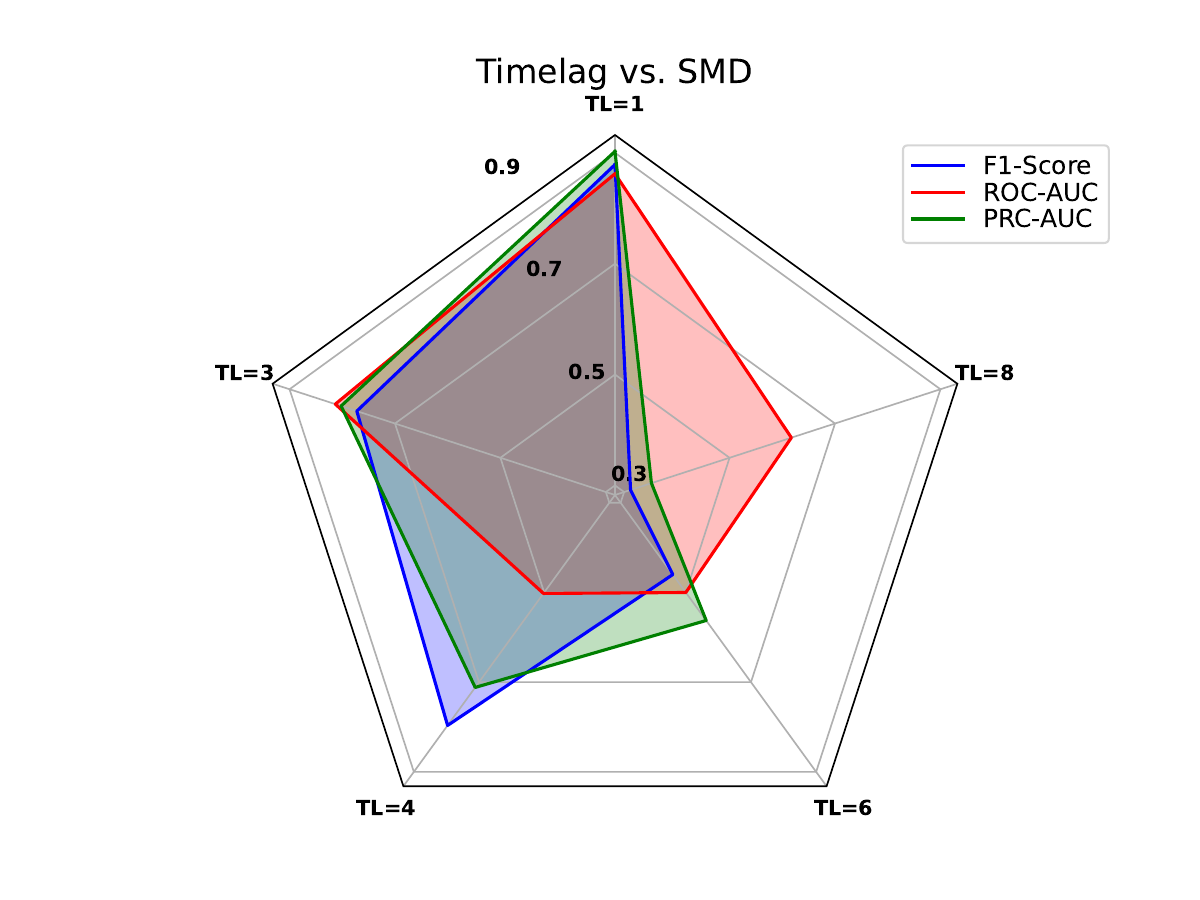}
}
\vspace{-0.3cm}
\caption{Performance analysis of INCADET on Timelag sensitivity.}
\label{timelag_analysis}
\end{figure*}

\begin{figure*}[htbp]
\centering
\subfigure[Performance vs. Batches: SWAT]{
\includegraphics[width=4.3cm, height=3.5cm]{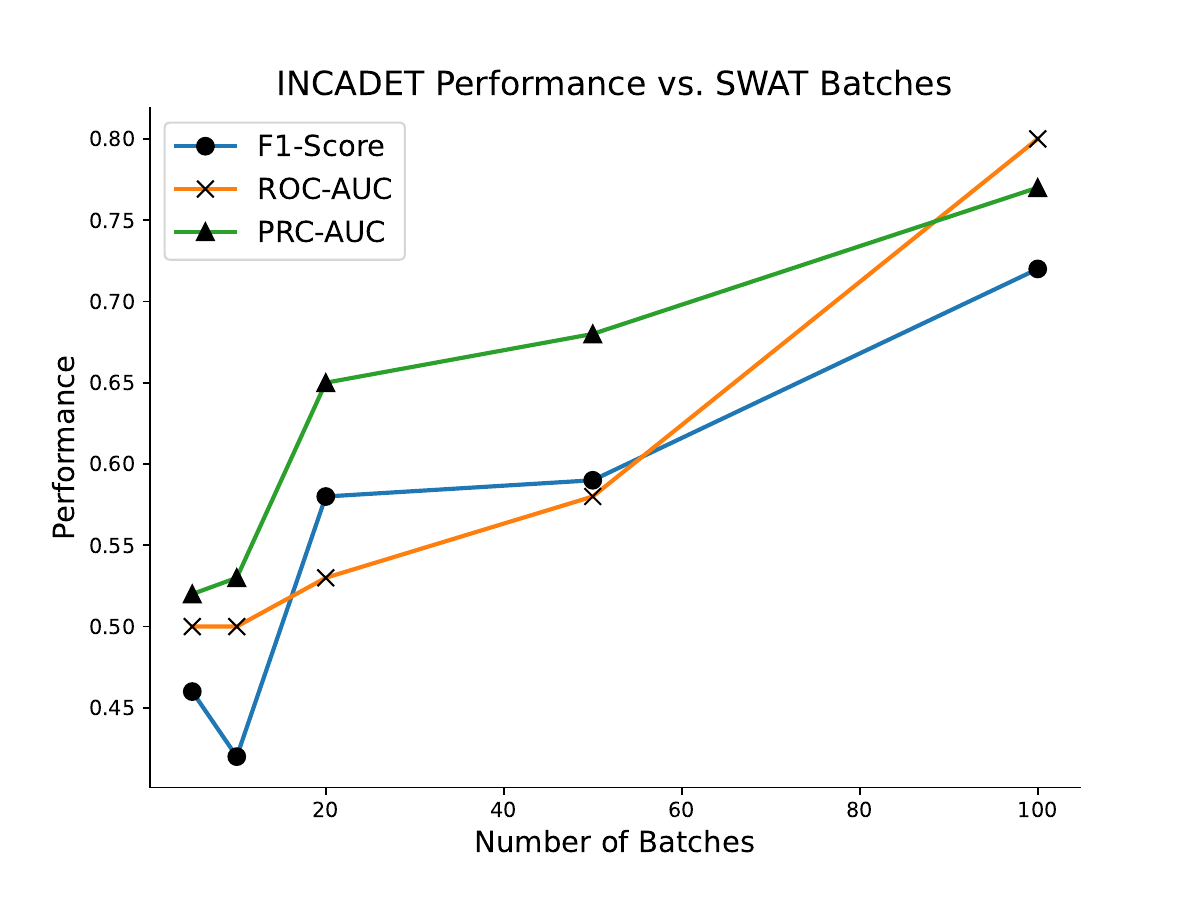}
}
\hspace{-3mm}
\subfigure[Performance vs. Batches: WADI]{ 
\includegraphics[width=4.3cm, height=3.5cm]{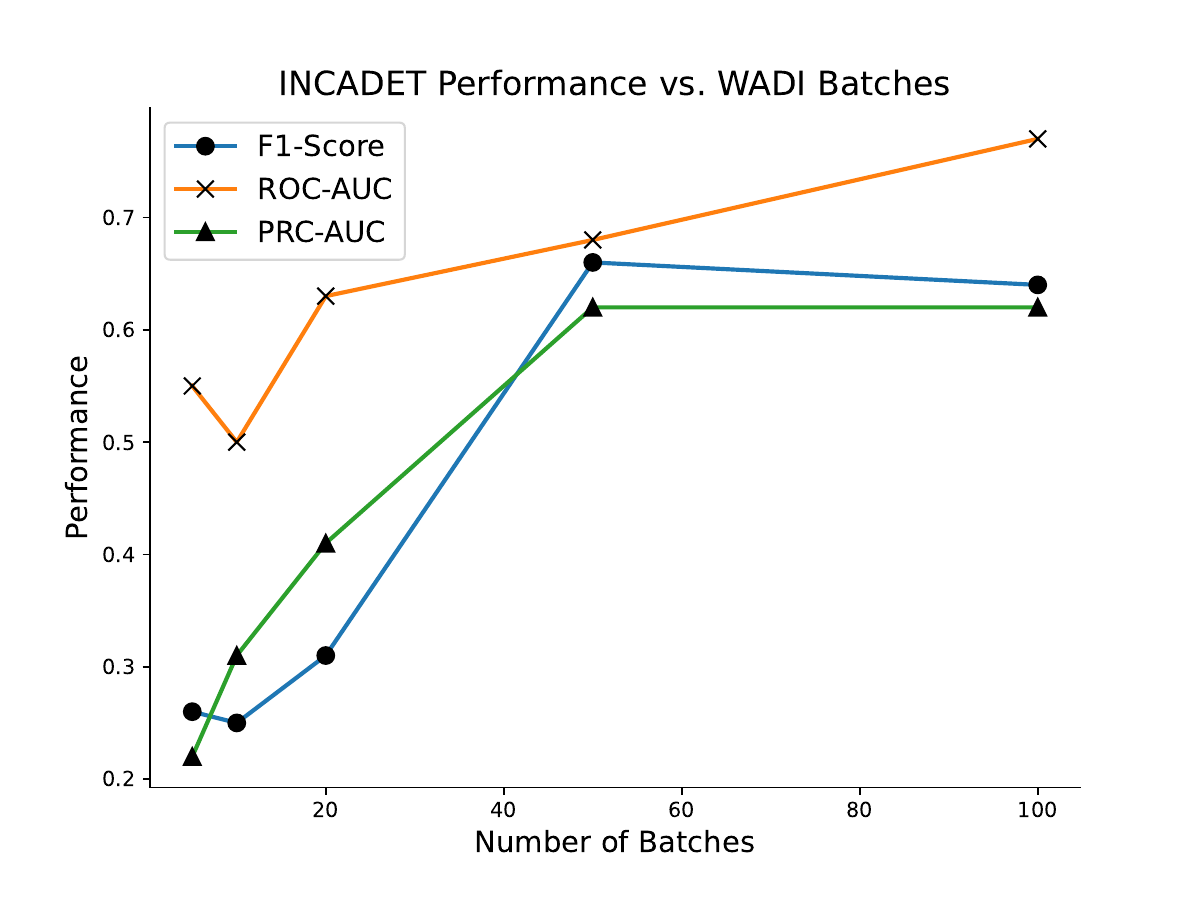}
}
\hspace{-3mm}
\subfigure[Performance vs. Batches: TE]{ 
\includegraphics[width=4.3cm, height=3.5cm]{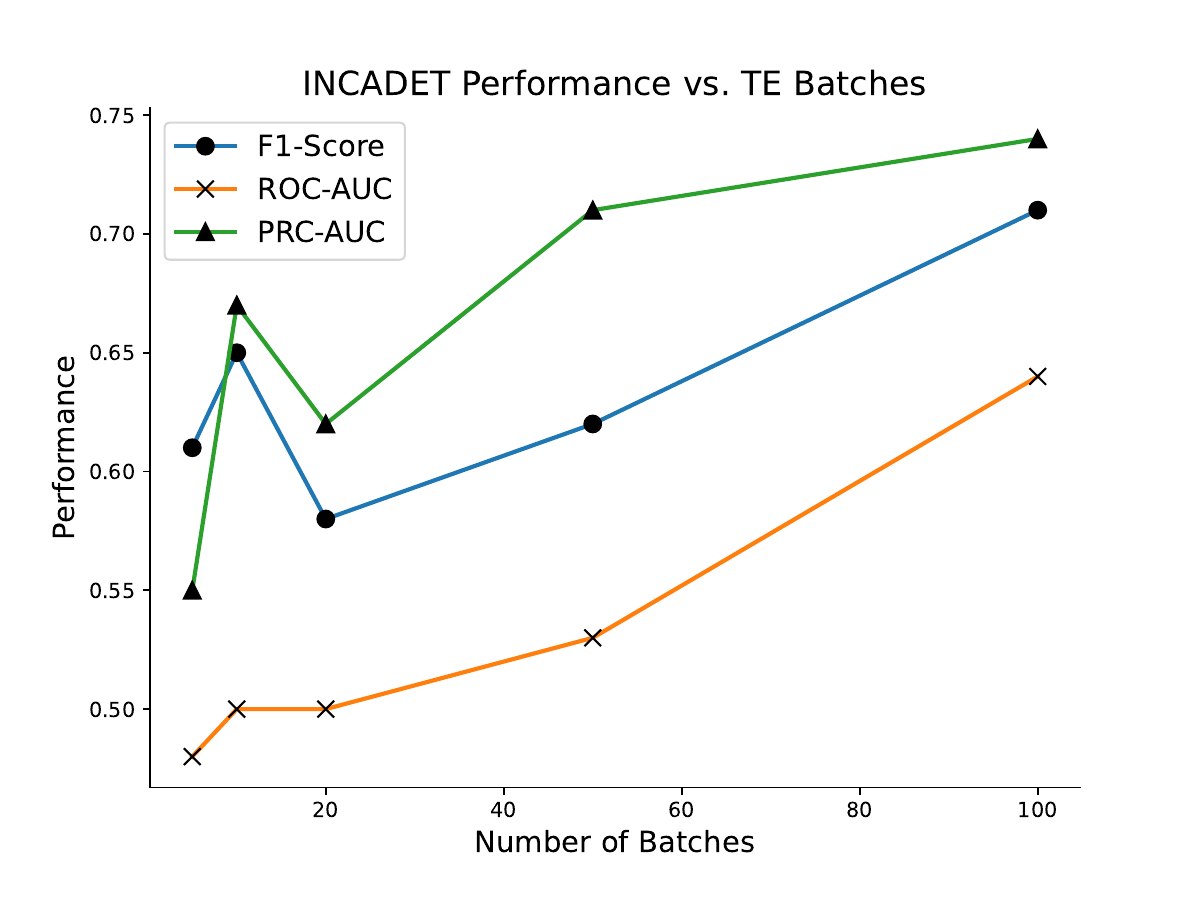}
}
\hspace{-3mm}
\subfigure[Performance vs. Batches: SMD]{ 
\includegraphics[width=4.3cm, height=3.5cm]{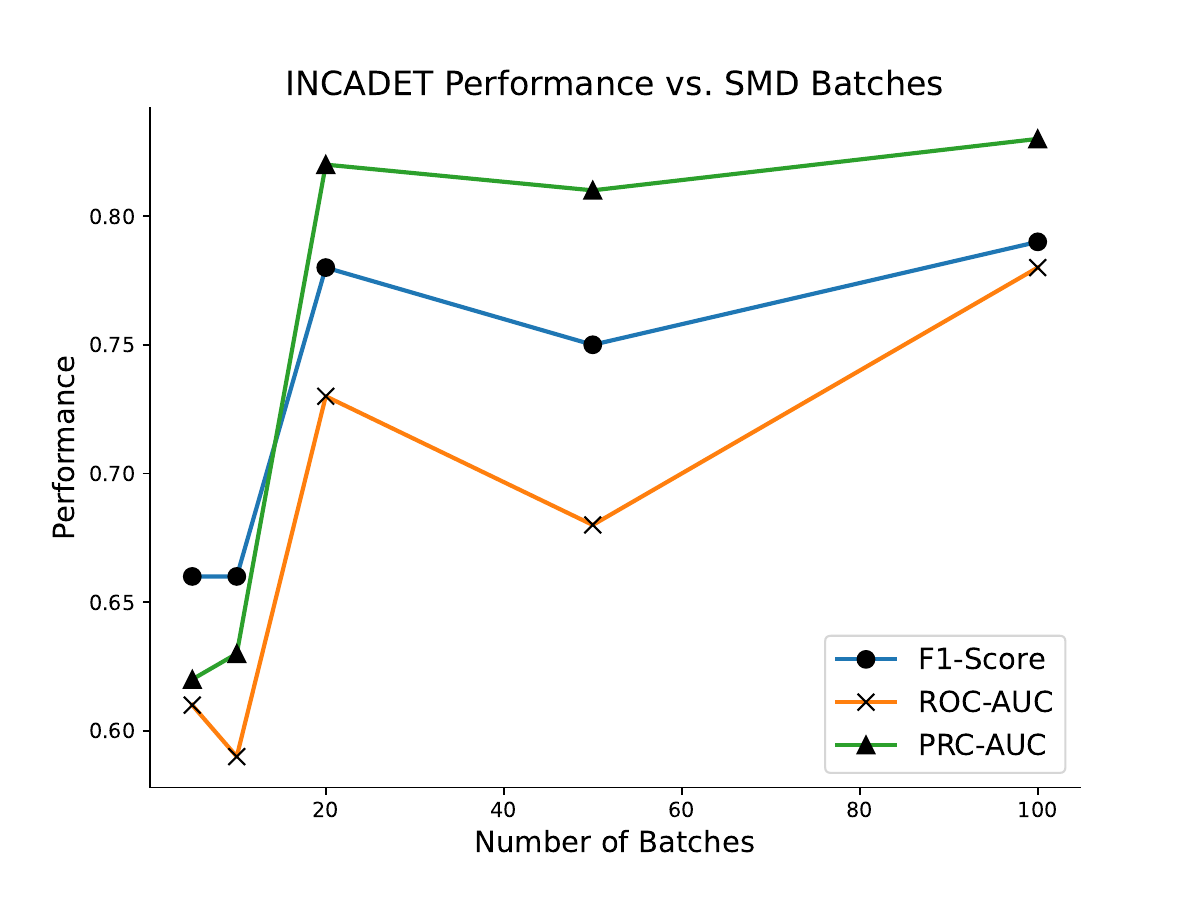}
}
\vspace{-0.3cm}
\caption{Performance analysis of INCADET on Scalability.}
\label{scalability_analysis}
\end{figure*}

\subsubsection{Efficiency Analysis: Offline vs. Incremental Causal Learning}
In this section, we compare the efficiency of the incremental causal graph-based frameworks over the offline models in terms of memory utilization and performance (\textbf{RQ5}). Fig. 5 shows the efficiency of the incremental framework over the offline sparsified causal graph framework in terms of Missed Alarm Rate (MAR) and structural hamming distance. Across real-world testbed datasets, the incremental framework leads to a significant drop in MAR ranging from $6.5\%$ to  $15.7\%$. This is possibly due to the ability to handle evolving temporal dependencies and distinguish spurious relationships from true causal relationships. To support this claim, a structural comparison of the causal graphs revealed a significant drop in the number of edges in the incremental frameworks, providing evidence of optimized memory usage compared to the offline models.

\subsubsection{Time Cost Analysis}
In this section, we compare the training times of our framework with various non-causal and causal anomaly detection models. We find that causal frameworks require less time than black-box graph-based methods, likely due to the efficiency of score-based causal graph learning techniques like DYNOTEARS, which operates with $\mathcal{O}(n)$ time complexity. Compared to offline causal graph models, our incremental framework trains faster, as the offline models rely on large historical datasets to build dense causal graphs, often with spurious edges. Additionally, our framework benefits from clearly defined convergence criteria, guided by causal domain expertise, which helps avoid spurious edges and overfitting, further reducing training time compared to baseline algorithms.

\begin{figure}[htbp]
\centering
\subfigure[SMV-CGAD vs. INCADET: MAR]{
\includegraphics[width=6cm, height=4cm]{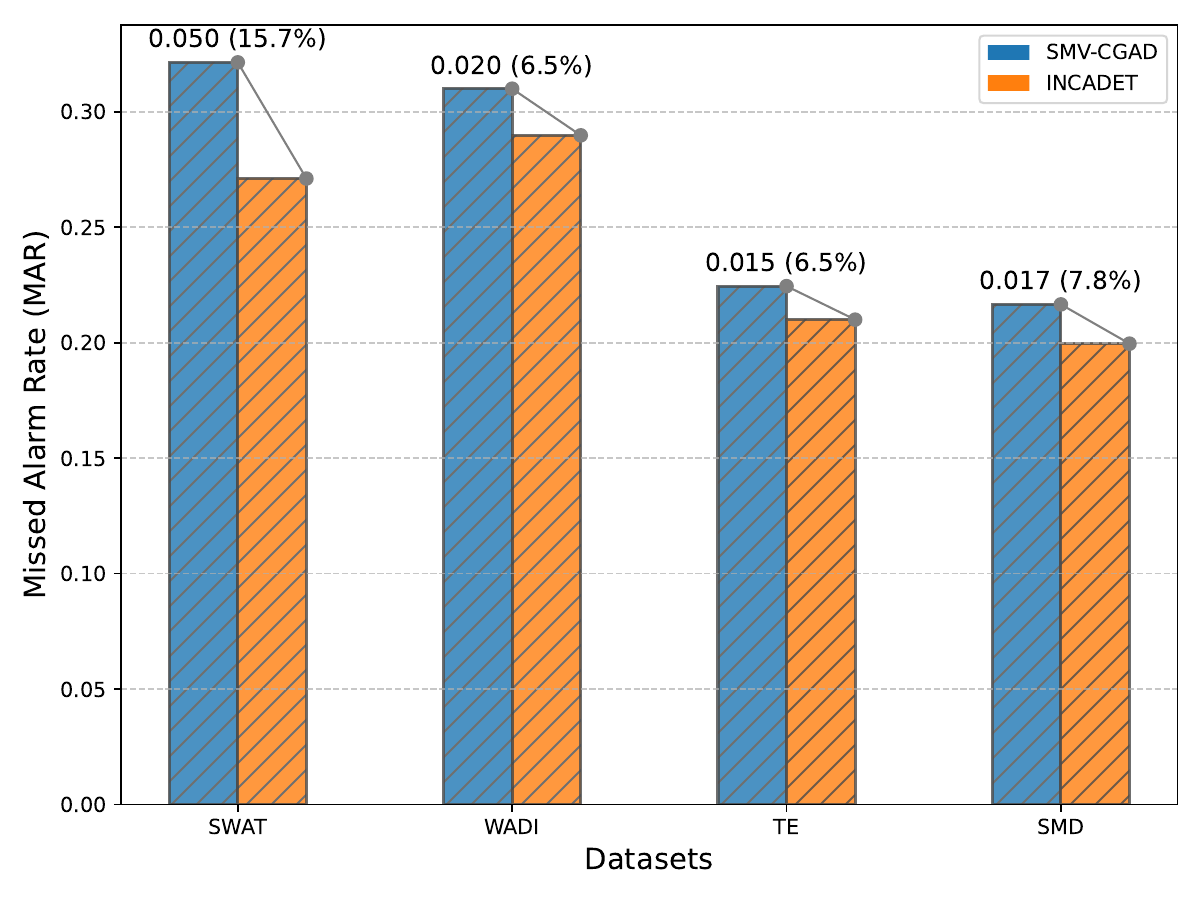}
}
\subfigure[SMV-CGAD vs. INCADET: \# Causal Edges]{ 
\includegraphics[width=6cm, height=4cm]{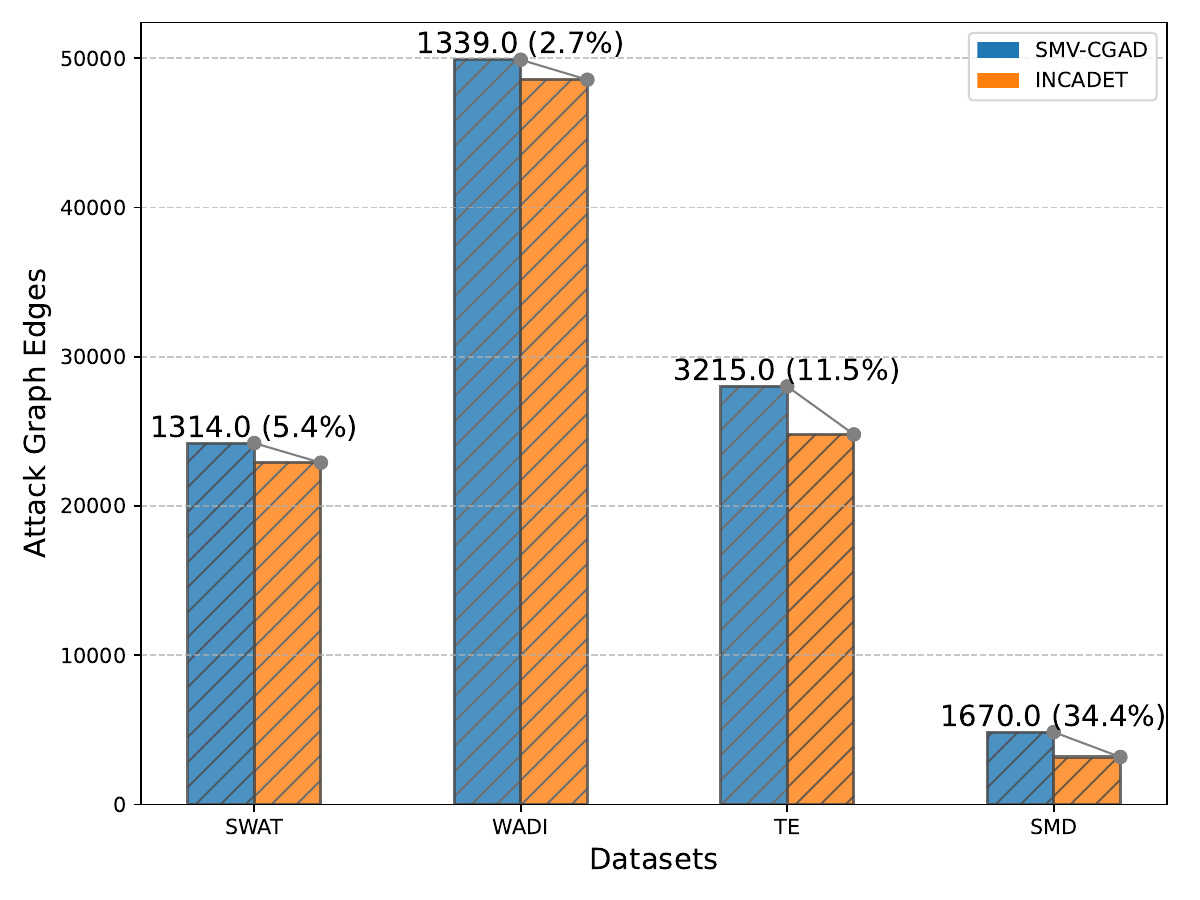}
}
\caption{Offline vs. Incremental Causal Frameworks}
\label{search}
\end{figure}

\section{Related Work}
\textbf{Anomaly Detection in Time Series Data} has been an evolving paradigm of Machine Learning research with the rise in sophistication of the attacks and its economic impact on domains like healthcare, finance, and public infrastructures \cite{6684530, Zamanzadeh_Darban_2024}. Evolving from traditional statistical-based approaches, the deep learning-based models \cite{9523565} like Autoencoders \cite{10.5555/3491440.3491613, Zhang_2019, lai2023contextaware} and Generative Adversarial Networks \cite{Li_2019, 9626552} showed prominence in effectively modeling and reconstructing the temporal dependencies and significant deviations can be identified as anomalies. The inherent relationships in the data are better captured as graphs and this led to the exploration of many graph-based approaches \cite{ho2023graph} to address temporal anomaly detection. These approaches have proved to be effective in handling complex temporal and spatial dependencies.

\textbf{Graph-based Anomaly Detection in Time-Series Data} captures anomalous events based on disrupted connectivity of the features from a wide spectrum of unstructured and multivariate data with temporal dependencies. The recent literature classifies these methods into 3 categories: Autoencoder-based \cite{tengDeepHypersphereRobust2018, 10.1145/3539597.3570371, LI2022101}, GAN-based \cite{10.1016/j.neunet.2023.07.026, Lu_2023, 9758699}, and Self-Supervised methods \cite{9599560, 10.1109/TKDE.2021.3119326, Caville_2022}. These categories of techniques can be utilized in both static and dynamic graph-based anomaly detection. However, the existing approaches suffer from spurious edges due to high-dimensional correlating variables and concept drift which has led to research in causality-based anomaly detection and root cause analysis \cite{10.1145/3627673.3680096, wang2023hierarchical, 10.1145/3580305.3599392, yang2022causal, chen2024semisupervised}. 

\textbf{Causal Learning and Inference in Time series} is a recent research area of interest across domains that solves the black-box interpretability and spurious correlation challenges of the traditional deep learning models. This unfolds the effects of intervention and interdependent causal relationships across time segments \cite{Moraffah_2021, ahmad2024deep}. Existing techniques are categorized into Granger-Causality \cite{make1010019, 9376668}, Constraint-based \cite{runge2020discovering}, Noise-based \cite{peters2012causal}, and Score-based \cite{wang2023hierarchical, pamfil2020dynotears} methods. Lately, these causal learning techniques have been proven to be effective in anomaly detection in multivariate time series data \cite{malarkkan2025rethinkingspatiotemporalanomalydetection, febrinanto2023entropy, heppel_et_al:OASIcs.DX.2024.11}. However, these methods assume the causal relationships to be predominantly static across time which does not hold true in real-time systems. In contrast, our work is an Incremental Causal Learning framework based on Causal Edge Reinforcement for Anomaly Detection. We utilize a score-based causal structure learning algorithm DYNOTEARS \cite{pamfil2020dynotears} and incrementally update the constructed graphs with unseen knowledge from the data stream.

\section{Conclusion and Future Scope}
In this paper, we propose a novel continual causal graph learning framework "INCADET" for cyberattack detection in critical public infrastructures. Our framework effectively captures the causal relationships among the different components like sensors and actuators in the infrastructure over a streaming data pipeline. The continual learning of the causal structure is achieved by leveraging causal domain knowledge of cyberattack events to sparsify the graph, preserve the spectral properties of the graph and storing this knowledge in a replay memory buffer for incrementally updating the graphs. To efficiently detect anomalies, a Deep Convolutional Graph Neural Network is employed, utilizing the learned causal graphs representing the system status as ground truth graphs. Addressing the shortcomings of existing anomaly detection frameworks and the research challenges in incremental learning, our framework demonstrates superior efficiency, interpretability, and scalability through extensive empirical evaluation on real-world datasets. However, there are certain challenges that needs to be addressed to deploy our framework in real world such as the causal knowledge accuracy, domain expert interruption bias, hyperparameter search-scalability trade-offs and theoretical guarantees for robust causal structure learning. This opens up more promising directions for future work, including exploring the use of large language models to harness the rich textual descriptions of causal knowledge within real-world infrastructures and integrate them with causal frameworks for cybersecurity systems.

\bibliographystyle{IEEEtran}
\bibliography{main}











\vfill

\end{document}